\RequirePackage{fix-cm}
\documentclass[smallextended]{svjour3}       
\smartqed  
\usepackage{graphicx}
\usepackage{multirow}
\usepackage{graphicx}
\usepackage[center]{caption}
\usepackage{subfigure}
\usepackage{amsmath}
\usepackage{mathtools}
\usepackage{listings}
\usepackage{color}
\usepackage{url}
\usepackage{tabularx,booktabs}
\usepackage{tikz}
\usepackage[backref, colorlinks=true, allcolors=blue, final]{hyperref}
\usepackage[sort&compress,numbers]{natbib}
\usepackage{etoolbox}
\begin{document}

\title{Responsive parallelized architecture for deploying deep
learning models in production environments\thanks{Nikhil Verma and Krishna Prasad are with Info Edge (India) Limited, Noida, India.}
}


\author{Nikhil Verma         \and
        Krishna Prasad 
}


\institute{Nikhil Verma  \at
              Info Edge (India) Limited\\
              Tel.: +91-6283273195\\
              \email{verma.nikhil@infoedge.com}\\
              Secondary \email{lih.verma@gmail.com}
}


\maketitle

\begin{abstract}
Recruiters can easily shortlist candidates for jobs via viewing their curriculum vitae (CV) document. Unstructured document CV beholds candidate's portfolio and named entities listing details. The main aim of this study is to design and propose a web oriented, highly responsive, computational pipeline that systematically predicts CV entities using hierarchically-refined label attention networks. Deep learning models specialized for named entity recognition were trained on large dataset to predict relevant fields. The article suggests an optimal strategy to use a number of deep learning models in parallel and predict in real time. We demonstrate selection of light weight micro web framework using Analytical Hierarchy Processing algorithm and focus on an approach useful to deploy large deep learning model-based pipelines in production ready environments using microservices. Deployed models and architecture proposed helped in parsing normal CV in less than 700 milliseconds for sequential flow of requests. 
\end{abstract}

\section{Introduction}

Process of decision making by humans is often prone to influences and societal biases which are drawbacks of cognitive thinking. Hiring candidates for jobs is one such affected area and companies are shifting to green human resource management (HRM) practices which are free from consequences of humanistic powers. Adoption of deep learning technologies to implement green HRM practices is a potentially fair approach to move into that direction. But unstructured information poses a huge problem for machines and algorithms to fetch relevant information. 

Curriculum-Vitae (CV) is a necessary document for applicants to show their avid behavior towards the open requirement. It is generally short but pithy and lists various skills, work-experience and educational profile of the applicant. Creativity, simplicity and imagination are some key features of CV writing and make it unstructured. Converting unstructured textual CV into structured data makes it useful at many places as in Applicant Tracking System (ATS), Applicant Ranking System (ARS) or CV recommender engines.


The unstructured CV is meant to contain information belonging to different categories mentioned in table \ref{table:CV_information_categories}. Deep learning techniques are helpful in determining relevant entities from raw CV text and propose output in a structured format. Once all the important fields required as per the business logic are collected, it becomes easy to apply search and filters for shortlisting candidates which may include preferred qualification details, matching skills or location of candidate at present.  
 
\begin{table}[h]
\scriptsize
\centering
\caption{CV information categories}
\label{table:CV_information_categories}
\begin{tabular}{| m{1.5em} | m{1.2cm}| m{5.6cm} |}\hline
&\textbf{Category}&\textbf{CV Fields}\\\hline
&  & Name, Date of birth, Mobile number, \\
1 & Personal & Email ID, Gender, Languages known,\\
& information & Address, City, Country, \\
& & Alternate mobile number, Alternate email ID\\\hline
2 & Skills & Key Skills\\\hline
& & Diploma (course, specialization),\\
3 & Education & Undergrd(course, specialization, institute, year),\\
& & Postgrd (course, specialization, institute, year), \\
& & Doctoral (course, specialization, institute, year)\\\hline
4 & Functional area & Functional area, Industry type, Role\\\hline
& & Current designation, Current employer,\\
5 & Work & Previous designation, Previous employer, \\
& experience & Salary, Total experience, Notice period, Experiences\\\hline
\end{tabular}
\end{table}


To seamlessly integrate and share output of deep learning models trained for CV Parser, developing application programming integrfaces (APIs) following representaional state transfer (REST) software architecture makes the objective highly achievable. An important step to make REST APIs highly responsive is the selection of web framework that is lightweight and has high throughput. A web framework range from full-stack to micro framework. Choosing a perfect framework via employing multi criteria decision making (MCDM) approach is one of the voluble point discussed in this article.  
 

From table \ref{table:CV_information_categories} it becomes evident that each category of information embedded in CV is highly specific and therfore require specialized deep learning models dedicated to extract fields belonging to a single category of CV information only. Deep learning models are both computation and time intensive in nature that surmount extra time required for processing the raw CV. Main aim of this article is to propose a parallelization based architecture, that could couple the time requirement of various sections, with no loss in output generated.

Once a responsive parallelized CV parser service has been built, it's important to ensure zero downtime of service to serve all the hours across the clock. This study suggests technique for deployment of service that uses deep learning models and was developed using lightweight micro web framework.  
 
\section{Related Litrature}
Selection of optimal web-framework from a large set of options has always been a challenging topic. In past automatic selection of web-frameworks by a software tool which use method same as to solve optimization problems is found helpful \cite{muhammed2020selecweb}. Standalone frameworks in isolation have been assessed in past as spring for J2EE Web development \cite{arthur2005spring} and also comparison of different frameworks has been done, illustratively comparing Django, Hadoop and eucalyptus \cite{rodriguez2010open} \cite{heitkotter2013comparison}. Various web development technologies as Python, PHP and Node.js have been compared \cite{lei2014performance} and Python is found to be suitable for building large scale websites.  

Artificial intelligence (AI) has been proposed as emerging technology for improvisation in the field of human resource management (HRM) \cite{devi2020role}. Deep Learning has been seen as promising solution to large and complex datasets \cite{najafabadi2015deep}. It has been suggested that fairness can be brought in field of recruitment by adopting green HRM practices which are AI-based and CV Parsing is one such practice\cite{ochmann12fairness} \cite{jia2020communication}.  

CVs with different mime type were converted to obtain text using Tika tool \cite{bucurusing}. Sections of text were segregated using section headings in \cite{sonar2012resume}. Bidirectional encoder representation transformer (BERT) is state-of-the-art model for obtaining embeddings that help in processing text \cite{bhatia2019end} \cite{annamoradnejad2020colbert} and has been used in past to achieve resume parsing and ranking job seekers based on their suitability to job description. Named entities from each section can be detected using Named Entity Recognition (NER) models where hierarchically-refined Label Attention Network (LAN) helps in finding long range label dependencies via leveraging label embeddings, making Bi-LSTM(LAN) outperforming Bi-LSTM (CRF or SoftMax) \cite{cui2019hierarchically}. Prediction-as-a-Service using distributed architecture has been used in natural language-based text mining service as a case study \cite{loreti2020parallelizing}.  

Industrial trend is to use deep learning solutions in form of service-oriented architecture \cite{briese2020towards} along with quick deployment and availability being major features of concern. Resume predictor is provided in form of a web-application \cite{amin2019best} that matches candidates to jobs for recruiters after parsing resume. Deep learning-based services are gaining high importance with field booming day by day \cite{ishakian2018serving} \cite{gujarati2017swayam}. Fast and responsiveness of such services is also equally important \cite{crankshaw2017clipper}. Usage of various techniques, tools along with challenges faced while developing scalable deep learning models using popular deep learning frameworks have been suggested in a brief study \cite{mayer2020scalable}. Nearly a dozen of open-source DL frameworks was investigated among which TensorFlow and Pytorch are found to have largest and second largest communities on git and stack overflow.  

Technique to design fault-tolerant web applications which are critical from business point of view had been discussed by author in \cite{rovnyagin2020distributed} focusing on cluster, load-balancer and backup services. Author in \cite{wolff2020enabling}, discussed cost of downtime of services in industrial setup. Architecture to implement RESTful APIs and usage of process management tool (Supervisor), has been in paper \cite{bogo2020component}. A survey suggests that microservice oriented architecture help in developing autonomous software components that help in isolating fine-grained business functionalities and has advantages as scalability, maintainability and independent deployment \cite{viggiato2018microservices}.  
 
\section{Materials and methods}
\subsection{Selecting highly responsive web framework} \label{SubSec:LightWeightWebFramework}
A web framework is a collection of modules which allow developers to write web applications without having to handle low-level details as protocols, sockets or process/thread management. There are many micro web-frameworks available in Python for web development. For detailed analysis with objective of selecting optimal framework for CV parser we studied three famous frameworks, selected after looking at their GitHub star and fork numbers.  
 
\begin{enumerate}
\item \textbf{Flask} \\
Github Star: 50.8k | Fork: 13.6k \cite{FlaskGithub}\\
Flask 1.1.1 is latest stable version of lighweight web server gateway interface(WSGI) based web-framework.  WSGI is Python standard described in PEP-3333 \cite{PEP3333}. It has an ablility to scale upto complex applications. After Django, which is a full-stack web-framework, it is second most famous choice of developers.

\item \textbf{FastApi} \\
Github Star: 15.4k | Fork: 1k \cite{FastApiGithub}\\
FastAPI is a modern, fast (high-performance), web framework for building APIs with Python 3.6+ based on standard Python type hints. Its stable version used for comparison in this article is 0.54.1. It has been used in past for IoT related services. 

\item \textbf{Falcon} \\
Github Star: 7.8k | Fork: 789 \cite{FalconGithub} \\
Latest version of Falcon is 2.0.0 used in this study. It is designed to support the demanding needs of large-scale microservices and responsive app backends. It is minimalist WSGI library in python for building speedy web APIs. It is used by growing number of organisations across the globe.


\end{enumerate}

\subsubsection{Test environment and hardware used}
Testbed consist of two programs running on same machine where one was web-client and other was the web-framework. Both of them were deployed on same machine so that impact of bandwidth during the test should not affect the overall selection mechanism. The machine used in this test runs Ubuntu 18.04.4 LTS, with an Intel(R) Core (TM) i5-8250U CPU @ 1.60GHz processor, 500GB disk space, DDR4 16GB RAM with speed 2400 MT/s.  

In addition, all non-essential processes on the machine were disabled to minimize the consumption of resources and kept fair in our test. Since our concern is to choose highly responsive web-framework for running deep learning based service, study demands performance testing of the frameworks being compared. For benchmarking, we use Apache Bench (Ab) tool of Apache \cite{ApacheBench}. Ab can make requests on local services to ensure that the time is just processing time, not including data transmission time on the internet/intranet or calculation time in local machine.  
 
\subsubsection{Test scenarios}
Following one factor at a time scheme, we made three test scenarios - "Hello world", "Finding value of Fibonacci", "File retrival from database".
\begin{itemize}
\item \textbf{Hello World}: This is one of the basic scenario for making a good web api.
\item \textbf{Finding value of Fibonacci}: This scenario is to test the framework under compute intensive i.e. CPU bound task, where value of Fibonacci series is calculated till its $100th$ term.
\item \textbf{File retrival from database}: This scenario is to test web-framework for a input-output (IO) intensive task, where a file is retrieved from NoSQL (MongoDB) storage and later stored on disk after retrival.
\end{itemize}

\subsubsection{Selection criteria} \label{SelectionCriteria}
Web-framework selection is an optimization problem where the goal is to select highly reponsive web-framework. We measured performance metrics for various alternatives that acted as our criteria for framework selection. Problem falls under category of multi criteria decision making approach and was solved via Analytical hierarchy processing (AHP) technique. Following is the list of criteria calculated using Ab tool for each alternative framework chosen in study.  
 
\begin{itemize}
\item Request per second: This is the number of requests served per second. This value is the result of dividing the number of requests served by the total time taken.
\item Time per request: The average time spent per request. 
\item Time taken for tests: This is the time taken from the moment the first socket connection is created to the moment the last response is received.
\item Time per concurrent request: The average time spent per concurrent request.
\item Total data transferred: The total number of bytes received from the server. This number is essentially the number of bytes sent over the wire.
\item Transfer rate: Total data transferred in bytes per unit time.
\end{itemize}

It is made sure that all requests are completed and there are no failure of requests.

\subsection{Parallelized architecture involving multiple deep learning models } 
A CV is a text document assembling knowledge of candidate from structurally different parts of one's portfolio. Each sentence fall under the major categories described in table \ref{table:CV_information_categories}.

\subsubsection{Text extraction}
CV document had no fixed format. Possible mimetypes of CV can be doc, docx, pdf, rtf, txt, odf, odt or else. Therfore extracting text from the document is primary step in process of unstructured text transformation. CV is first parsed through Apache Tika (version 1.24) \cite{ApacheTika} which can extract text and metadata from more than thousand different file types. Plain text extracted using tika was cleaned before feeding to any model.

\subsubsection{Text segmentation}
Each sentence of the CV text extracted using tika was then categorized to belong to one of the following class:-
\begin{enumerate}
\item Personal Information
\item Education
\item Work Experience
\item Others
\end{enumerate} 
The classifier(Sequential model) used to segment sentences was trained on manually annotated CVs. For training, sentence embedding was obtained using state-of-the-art BERT encoder pretarained $uncased_L-12_H-768_A-12$ english model \cite{devlin2018bert}. For each input sentence, the BERT Encoder calculates a 768 units long vector which is input to the classifier. Classifier summary is:-
\begin{lstlisting}
Model: "sequential_1"
____________________________________________
Layer     (type)    Output Shape    Param #   
============================================
dense_1   (Dense)   (None, 200)      153800 
____________________________________________
dense_2   (Dense)   (None, 4)         804       
============================================
Total params: 154,604
Trainable params: 154,604
Non-trainable params: 0
\end{lstlisting}
Once sentences have been tagged using the classification scheme, they become fit for the next step in process.

\subsubsection{Model training, storage and prediction}
Tagged sentences belonging to each particular section can be used to train deep learning based Named Entity Recognition (NER) models using Bi-LSTM technique \cite{cui2019hierarchically}. Nearly 50000 resumes were tagged using manual tagging technique for each field/named entity of various sections to train the system. Once NER models have been learnt, now they are fit for prediction.

With abundant data learnt via deep learning models, model file size also grows and it becomes challenging to serialize such large deep learning models. For privately storing large models, centralized document storage using Mongo-DB Grid-FS API which divides any file to chunks for storage, was used. 

\begin{figure}[ht]
\centering
\includegraphics[scale=0.30]{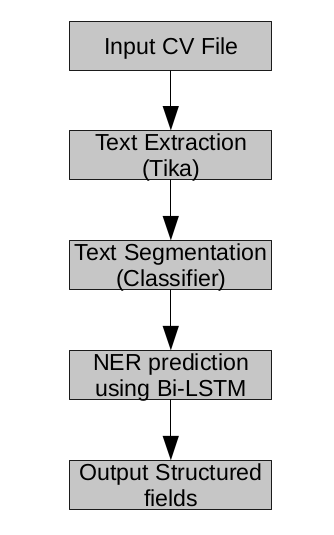}
\caption{Flow for NER prediction}
\label{fig:FlowForNERPrediction}
\end{figure}

Once training and storage of model is completed, the ground is set for prediction of named entities from different sections. Steps to follow for prediction via single NER model are described in fig \ref{fig:FlowForNERPrediction}. 

\subsubsection{Parallelization}
\begin{figure}[ht]
\centering
\includegraphics[scale=0.30]{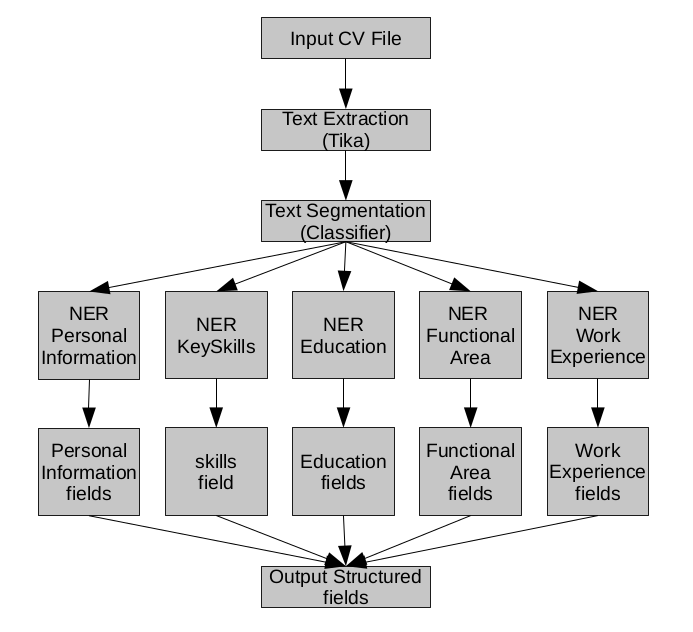}
\caption{Flow for NER prediction using multiple Bi-LSTM based NERs}
\label{fig:FlowForNERPredictionUsingMultipleBi-LSTMBasedNERs} 
\end{figure}

Prediction of named entities from one model is independent of entities obtained from other models. Therefore, after obtaining sectioning results, each section was fed to corresponding model and processed in parallel using multiprocessing standard library of python. For this study Process module in multiprocessing library of python was used for parallelly processing sections. Fig \ref{fig:FlowForNERPredictionUsingMultipleBi-LSTMBasedNERs} shows an overview of proposed mechanism for parallel processing.  
 
\subsection{Deployment of multiple specialized services}
A pre-requirement for obtaining named entities for various sections using specifically trained Bi-LSTM models used for prediction as shown in fig \ref{fig:FlowForNERPredictionUsingMultipleBi-LSTMBasedNERs} is availability of loaded model which are ready for predicting on any next text sent to it. This can be achieved by deploying each NER as a microservice that is predicting on a specific set of sentences, better summarized as Prediction-as-a-Service (PaaS), follows RESTful architecture and communicate using standardized interfaces and lightweight protocol. Downtime of any independent PaaS API used for each different section of fields will directly affect the overall accuracy of CV Parsing process.  

\begin{figure}[ht]
\centering
\includegraphics[scale=0.22]{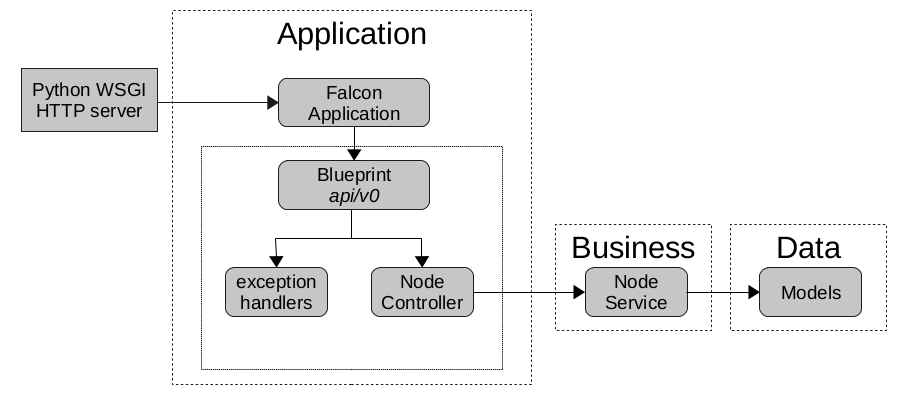}
\caption{Prototype implementation of RESTful API}
\label{fig:PrototypeImplementationOfRESTfulAPI}
\end{figure}

Using Web API design paradigm, it's easy to break API into three logical layers namely Application layer, Business layer and Data layer as shown in fig \ref{fig:PrototypeImplementationOfRESTfulAPI}. The Application layer is a place where the controllers are used for translating HTTP incoming requests and outgoing responses, and for validating payloads after encoding and decoding them. The Business layer is where the business logic of the API resides, with business rules and workflows defined to suitably interact with the Data layer. The Data layer is used for storage of all static information required in the process. 

Described ahead is deployment strategy designed to get rid of downtime problem:-

\subsubsection{Nginx and supervisor}
The design architecture proposed involve nearly half-a-dozen RESTful APIs, that need to be started in manner such that pre-requisite APIs are always working before the main CV Parser API is deployed. This signifies a priority in deployment of APIs. The solution is to put to use process management system as open-source Supervisor. Supervisor gives access to users to manage multiple processes on Unix-based operating systems that are based on client-server model where the server called as supervisor demon (supervisord) is responsible for starting subprocesses on demand. Clients can ask supervisord to spawn subprocesses through CLI known as supervisorctl. Enhancing supervisord is easy via configuration file commonly known as supervisor.conf. All commands to start APIs were forementioned in this configuration file along with priority keyword defined. Each API was deployed on more than one (actually three) machines.

Once all PaaS APIs are up and started running on different machines offering high availability (HA), they were up streamed to a single endpoint using NGINX upstreaming feature. NGINX is event-based (asynchronous), non-blocking I/O architecture-based http server. Exploiting usage of NGINX as load balancer, work load was distributed among multiple computer resources which in our case are PaaS API end points making a cluster. Out of three APIs for each independent PaaS, one of them acted as a backup service and other replicas are serving request in a round robin fashion.  
 

\section{Methodology} \label{Methodology}

\subsection{Analytical Hierarchy Processing(AHP)}\label{Analytical Hierarchy Processing(AHP)}
\begin{enumerate}
\item \textbf{Hitting framework}: Total of 10000 requests with 30 concurenecy hitting one framework at a time were run. Ab tool results were jotted to be sent to AHP as input data.

\item \textbf{Defining AHP graph}: Following are notable components of AHP graph:-
  \begin{itemize}
	\item \textbf{Goal}: Selection of light weight, highly responsive, micro web-framework is the goal of optimization problem at hand.
	\item \textbf{Criteria}: Various criteria for pairwise comparison are mentioned in section \ref{SelectionCriteria}.
	\item \textbf{Alternatives}: Flask, FastApi and Falcon are three alternatives to achieve the goal.
  \end{itemize}
The AHP graph proposed for the problem in study is shown in fig \ref{fig:Analytical Hierrachy Processing Graph}.

\begin{figure*}
\centering
\includegraphics[scale=0.30]{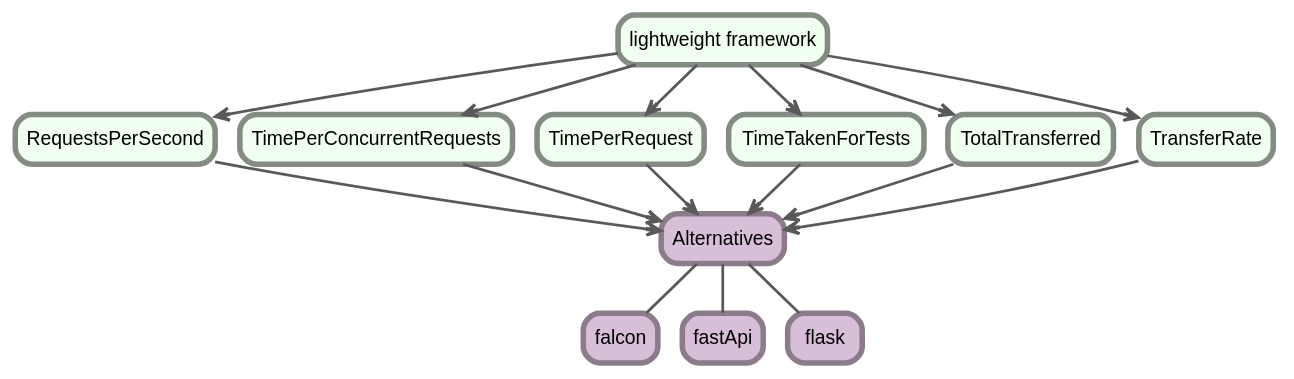}
\caption{Analytical Hierrachy Processing Graph}
\label{fig:Analytical Hierrachy Processing Graph} 
\end{figure*}

\item \textbf{Pairwise comparison}: AHP do pairwise comparison of alternatives w.r.t criteria. One need to define pairwise preference of alternatives or criteria in form:-
\begin{equation}
- [option1, option2, pairwise preference]
\end{equation}
where "pairwise preference" is a real number in range [1/9, 9]. The number in fraction with larger numerator implies option1 is prefereable than option2 and in case denominator is larger than vice-versa happens. 

From the list of criteria selected for study, criteria related to time (Time per concurrent request, Time per request and Time taken for tests) are the one for which we need to prefer the alternative having smaller numeric value, therefore the AHP logic used for the same is:-
\begin{lstlisting}
criteria:
  children: *alternatives
  preferences:
    pairwiseFunction: function(a1, a2) min(9, max(1/9,
			a2$criteria/a1$criteria))
\end{lstlisting}

For other criteria (Throughput, Transfer rate and Total data transferred) the parwise function is defined as:-
\begin{lstlisting}
criteria:
  children: *alternatives
  preferences:
    pairwiseFunction: function(a1, a2) min(9, max(1/9,
			a1$criteria/a2$criteria))
\end{lstlisting}

No criteria was prefered against other as parwise comparison of criteria was defined as:-
\begin{equation}
- [criteria1, criteria2, 1]
\end{equation}

\end{enumerate} 
The code is select highly responsive web framework along with running AHP on AB test results is uploaded here \footnote[1]{\url{https://github.com/lihkinVerma/Selecting-Highly-responsive-Webframework}}.

\subsection{Parallelizing web-services}
\begin{enumerate}
\item \textbf{Developing Independent microservices} After selecting optimal web-framework, microservice was developed for each model, that retrieve model from database and then load it for prediction. Microservices developed using RESTful architecture, following Prediction-As-A-Service (PaaS) paradigm communicate information in standard json format. Each RESTful PaaS had an API endpoint which on a POST call with predefined payload format, containing headers and data, predict structured fields of a section.

\item \textbf{Sentence classification} Once all independent API end points have started working, main CV Parser PaaS was deployed. It is responsible for consuming raw CV file as input. Text was extracted from this file using tika server, deployed as an independent service and after doing cleaning, it was sent to BERT for finding text embeddings. Sectioning classifier then tagged each sentence of the text.  
\item \textbf{Using parallelized architecture} Tagged sentences were sent to respective PaaS API end points in following manner: - 
\begin{enumerate} 
\item Personal Information section, sent to Personal Information PaaS. 
\item Education section, sent to Education PaaS. 
\item Work Experience section, sent to Work Experience PaaS. 
\item Work Experience and Others section, sent to Skills PaaS. 
\item Others section, sent to Functional Area PaaS. 
\end{enumerate} 

This is achieved in parallelized fashion since prediction of fields belonging to one section are independent of fields belonging to the other. Proposed architecture for parallelism is shown in fig \ref{fig:Parallelized architecture involving multiple deep learning service}.

Request reaching CV Parser PaaS, will first reach web-framework application which define URL routes. Request reaching file parser route will be checked for POST data to have raw file and valid headers. Raw CV file is then passed to CV Parser Service that interacts with Sectioning service and obtain four sections. Parallel request is then passed to independent PaaS in forementioned style. Results from all microservices are combined and a full-fledged structured output is then returned to CV Parser service which communicates it further and finally back to the caller api endpoint.  
 
\begin{figure*}
\centering
\includegraphics[scale=0.40]{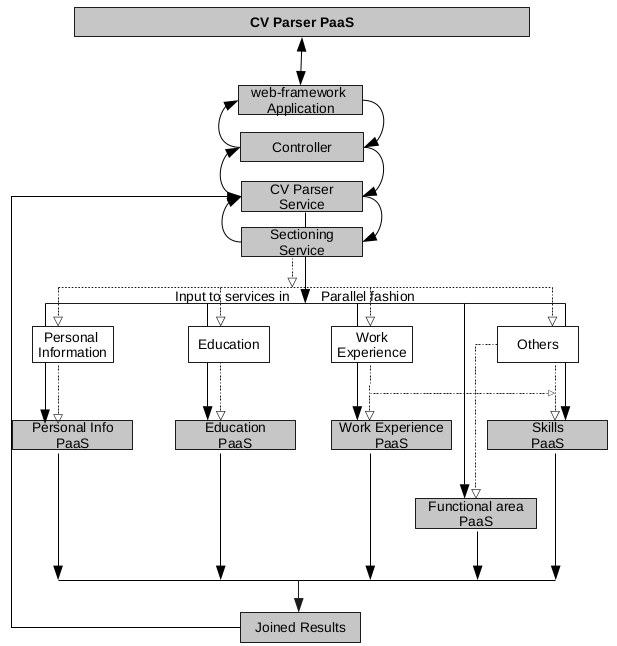}
\caption{Parallelized architecture involving multiple deep learning service.\\(Here
"dotted" line represents logical flow of data while "solid" line represents actual request flow.)
}
\label{fig:Parallelized architecture involving multiple deep learning service} 
\end{figure*}

\end{enumerate}

\subsection{Deploying services} \label{Independent services}
Independent PaaS services were defined in priority fashion in supervisor.conf file in below described order:-
\begin{itemize}
\item \textbf{Priority 0}: Tika server is deployed.
\item \textbf{Priority 1}: BERT server is started, for BERT client; to be used in CV Parser PaaS for text embedding.
\item \textbf{Priority 2}: Indpendent section PaaS required for predicting particular section fields are exposed for CV Parser PaaS. It includes following API end points:-
\begin{itemize}
\item Personal Information PaaS
\item Education PaaS
\item Work Experience PaaS
\item Skills PaaS
\item Functional Area PaaS
\end{itemize} 
\item \textbf{Priority 3}: Finally, CV Parser Service is deployed for converting unstructured CV files to structured JSON fields. 
\end{itemize}

Each PaaS described above was started on three machines for HA, making a cluster which were then upstreamed to a single uri using following illustrative entry of NGINX configuration file.

\begin{lstlisting}
upstream parser-independent-PaaS {
	keepalive 10;
	server ipAddress1:Port1 max_fails=3 fail_timeout=15s;
	server ipAddress2:Port2 max_fails=3 fail_timeout=15s;
	server ipAddress2:Port3 backup;
	}	
server {
    listen portUpstream;
    server_name parser-independent-PaaS.host;
    location / {
        proxy_set_header Host $http_host;
        proxy_http_version 1.1;
        proxy_pass http://parser-independent-PaaS;
    }
}
\end{lstlisting}

\section{Evaluation and Results} \label{Sec:Results}

\subsection{Analysis of AHP results}
Ab tool results which are metrics used in study for micro web-framework selection are mentioned in table \ref{ApacheBenchToolResults}. Applying the methodology proposed in section \ref{Analytical Hierarchy Processing(AHP)}, for each scenario the AHP results obtained are mentioned in tables \ref{AHP results for Hello World}, \ref{AHP results for "Finding value of Fibonacci"} and \ref{AHP results for "File retrival from database"}.

For the basic scenario of "Hello world", Falcon is decisively first out of three frameworks, selected in study, with 50.5\% as the percentage of selection to achieve the goal, followed by FastApi having 31.7\% and Flask is ranked third for this test scenario whose selection percentage is 17.8\%.  
For compute-intensive test scenario "Finding value of Fibonacci", Falcon batches first position with percentage share of 49.1\%. Next in the race is framework with 33.0\% selection chances namely FastApi, while Flask has 17.9\% for this test scenario.

For IO-intensive test scenario "File Retrieval from Database", frameworks are ranked as Falcon, Flask and FastApi, whose selection percentages are 34.1\%, 33.2\% and 32.7\% respectively. Unlike for basic and compute-intensive test scenarios, here FastApi and Flask frameworks interchange their position but Falcon remains the top choice for selection. Contribution of each criteria to determine total percentage for each alternative is mentioned in tables \ref{AHP results for Hello World}, \ref{AHP results for "Finding value of Fibonacci"} and  \ref{AHP results for "File retrival from database"}.

Observing the results for all three test scenarios, Falcon is the best of all the alternatives selected in study. It is micro web-framework, light weight and also highly responsive for developing service-oriented architecture, involving CPU bound or IO bound computations. 

\newcolumntype{C}{>{\centering\arraybackslash}X} 
\setlength{\extrarowheight}{1pt}
\begin{table*}[h]
 \caption{Apache Bench tool results}
\label{ApacheBenchToolResults}
\begin{tabularx}{\textwidth}{@{}l*{7}{C}c}
\toprule
Test Scenario     & Framework & Time per concurrent request(ms) & Requests per second & Time per request (ms) & Transfer rate (Kbps received) & Total transferred (bytes) & Time taken for tests (s) \\ 
\midrule
 & Falcon & 23 & 4274 & 4 & 680 & 1630000 & 2\\
Hello World & FastApi & 37 & 2650 & 7 & 357 & 1380000 & 3 \\
 & Flask & 84 & 1180 & 16 & 190 & 1650000 & 8 \\
\addlinespace
Finding value  & Falcon & 25 & 3969 & 5 & 610 & 1730000 & 2 \\
of Fibonacci  & FastApi & 38 & 2579 & 7 & 372 & 1480000 & 3  \\
 & Flask & 88 & 1126 & 17 & 192 & 1750000 & 8 \\
\addlinespace
File retrival  & Falcon & 701 & 142 & 140 & 22 & 1600000 & 70 \\
from database & FastApi & 693 & 144 & 138 & 19 & 1360000 & 69 \\
 & Flask & 729 & 137 & 145 & 21 & 1620000 & 72 \\
\bottomrule
\end{tabularx}
\end{table*}

\begin{table*}[h]
\caption{AHP results for "Hello World"}
\label{AHP results for Hello World}
\begin{tabularx}{\textwidth}{@{}l*{4}{C}c}
\toprule
 &  Weight & Falcon & FastApi & Flask\\ 
\midrule
Light framework & 100\% & 50.5\% & 31.7\% & 17.8\% \\\hline
Time per concurrent request & 16.7\% & 8.7\% & 5.8\% & 2.2\%\\
Requests per second &  16.7\% & 8.8\% & 5.4\% & 2.2\%\\
Time per request &  16.7\% & 8.8\% & 5.5\% & 2.4\%\\
Transfer rate &  16.7\% & 9.2\% & 5.2\% & 2.3\%\\
Total transferred & 16.7\% & 9.2\% & 4.8\% & 2.6\%\\ 
Time taken for tests & 16.7\% & 5.8\% & 4.9\% & 5.9\%\\
\addlinespace

\bottomrule
\end{tabularx}
\end{table*}

\begin{table*}[h]
\caption{AHP results for "Finding value of Fibonacci"}
\label{AHP results for "Finding value of Fibonacci"}
\begin{tabularx}{\textwidth}{@{}l*{4}{C}c}
\toprule
 &  Weight & Falcon & FastApi & Flask\\ 
\midrule
Light framework & 100\% & 49.1\% & 33.0\% & 17.9\% \\\hline
Time per concurrent request & 16.7\% & 8.7\% & 5.8\% & 2.2\%\\
Requests per second &  16.7\% & 8.6\% & 5.6\% & 2.4\%\\
Time per request &  16.7\% & 8.6\% & 5.6\% & 2.4\%\\
Transfer rate &  16.7\% & 8.3\% & 5.9\% & 2.4\%\\
Total transferred & 16.7\% & 9.0\% & 5.0\% & 2.6\%\\ 
Time taken for tests & 16.7\% & 5.8\% & 5.0\% & 5.9\%\\

\bottomrule
\end{tabularx}
\end{table*}

\begin{table*}[h]
\caption{AHP results for "File retrival from database"}
\label{AHP results for "File retrival from database"}
\begin{tabularx}{\textwidth}{@{}l*{4}{C}c}
\toprule
 &  Weight & Falcon & Flask & FastApi\\ 
\midrule
Light framework & 100\% & 34.1\% & 33.2\% & 32.7\% \\\hline
Time per concurrent request & 16.7\% & 5.6\% & 5.4\% & 5.7\%\\
Requests per second &  16.7\% & 5.6\% & 5.4\% & 5.7\%\\
Time per request &  16.7\% & 5.6\% & 5.4\% & 5.7\%\\
Transfer rate &  16.7\% & 5.6\% & 5.4\% & 5.7\%\\
Total transferred & 16.7\% & 5.9\% & 5.6\% & 5.1\%\\ 
Time taken for tests & 16.7\% & 5.8\% & 5.9\% & 4.9\%\\
\bottomrule
\end{tabularx}
\end{table*}

\subsection{Analysis of proposed parallelized architecture for deep learning based web-services}
\begin{figure}[ht]
\centering
\includegraphics[scale=0.25]{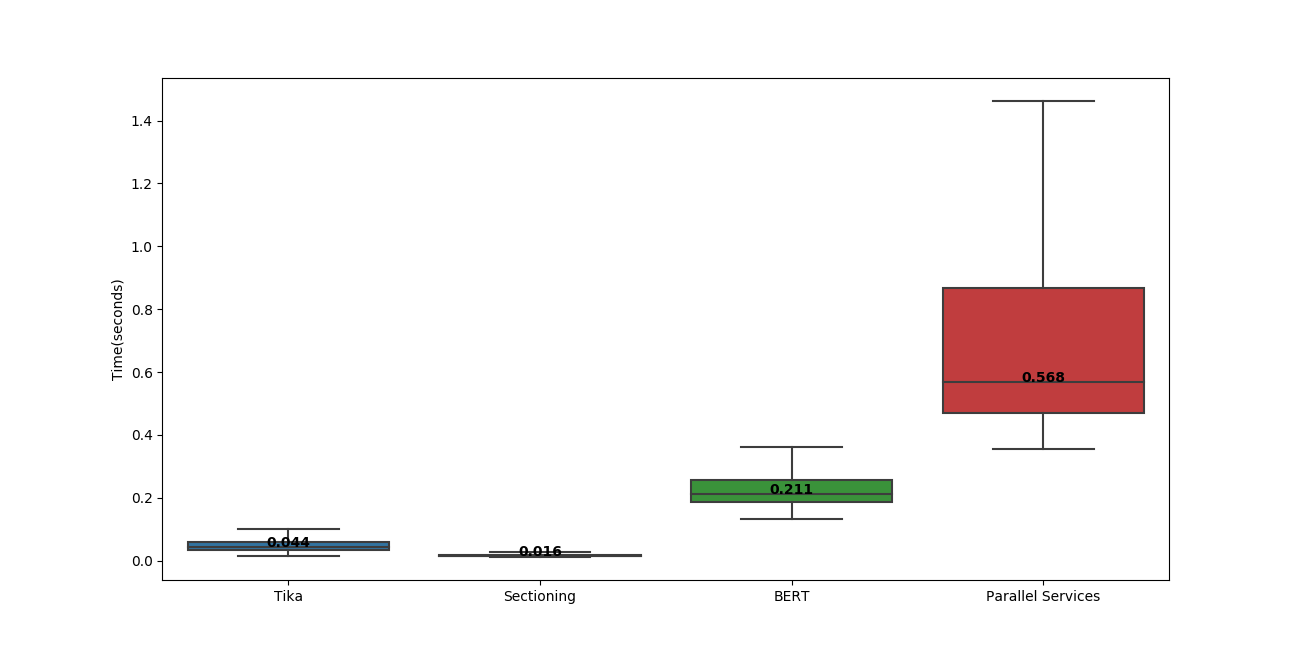}
\caption{Time taken by Tika, Sectioning classifier, BERT and Parallel serices in CV Parser PaaS}
\label{fig:totalTimeAnanlysis}
\end{figure}

The set of 1500 CVs (containing 500 CVs each of type doc, docx and pdf) of varying length were sent to the CV Parser PaaS and logs obtained were then analyzed for finding time taken at each step by the parser. Table \ref{Statistics for time taken at each step by CV Parser PaaS(in seconds)} shows statistics analyzed for Tika, Sectioning classifier, BERT encoding and getting joined results from parallel services, whose boxplots are shown in fig \ref{fig:totalTimeAnanlysis}. Fig \ref{fig:totalTimeAnanlysis} shows that calling independent PaaS APIs in parallel is most time-consuming step followed by generating text embedding from the BERT, tika text extraction and using sectioning classifier in that order. 

\begin{table*}[h]
\caption{Statistics for time taken at each step by CV Parser PaaS(in seconds)}
\label{Statistics for time taken at each step by CV Parser PaaS(in seconds)}
\begin{tabularx}{\textwidth}{@{}l*{5}{C}c}
\toprule
 &Tika  &Sectioning      &BERT &Parallel Services\\ 
\midrule
mean      &0.052       &0.017     &0.227              &0.767\\
std      &     0.037 &       0.004 &     0.059 &              0.489\\
min      &     0.016 &       0.010 &     0.131 &              0.356\\
25\%       &     0.033 &       0.014 &     0.185 &              0.470\\
50\%       &     0.044 &       0.016 &     0.211 &              0.568 \\
75\%       &     0.060 &       0.019 &     0.255 &              0.867\\
max       &     0.485 &       0.053 &     0.596 &              4.097\\
\bottomrule
\end{tabularx}
\end{table*}

\begin{figure}[ht]
\centering
\includegraphics[scale=0.25]{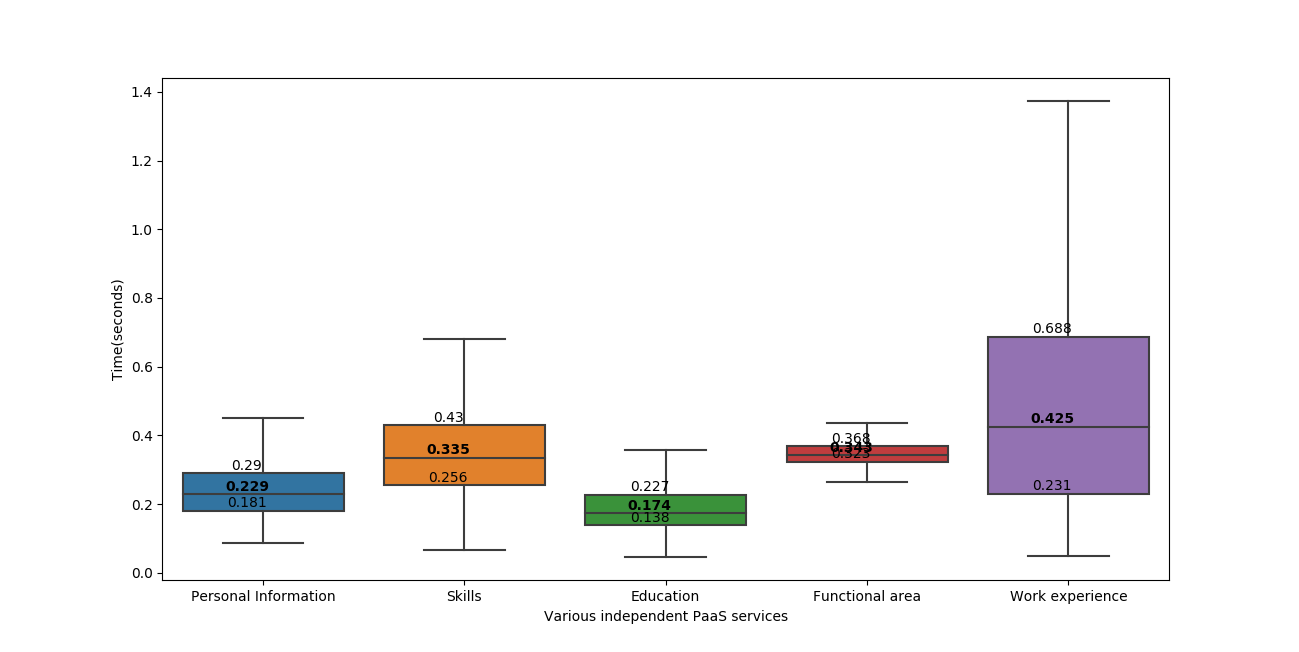}
\caption{Time taken by each Indpendent PaaS}
\label{fig:totalTimeAnanlysisforParallelPaaS}
\end{figure}

Each independent service mentioned in section \ref{Independent services} takes different amount of time. Time taken by each PaaS called while Parallel processing in CV Parser PaaS is shown in fig \ref{fig:totalTimeAnanlysisforParallelPaaS}. Observable is that work experience is most time consuming PaaS out of all.

\begin{figure}[ht]
\centering
\includegraphics[scale=0.25]{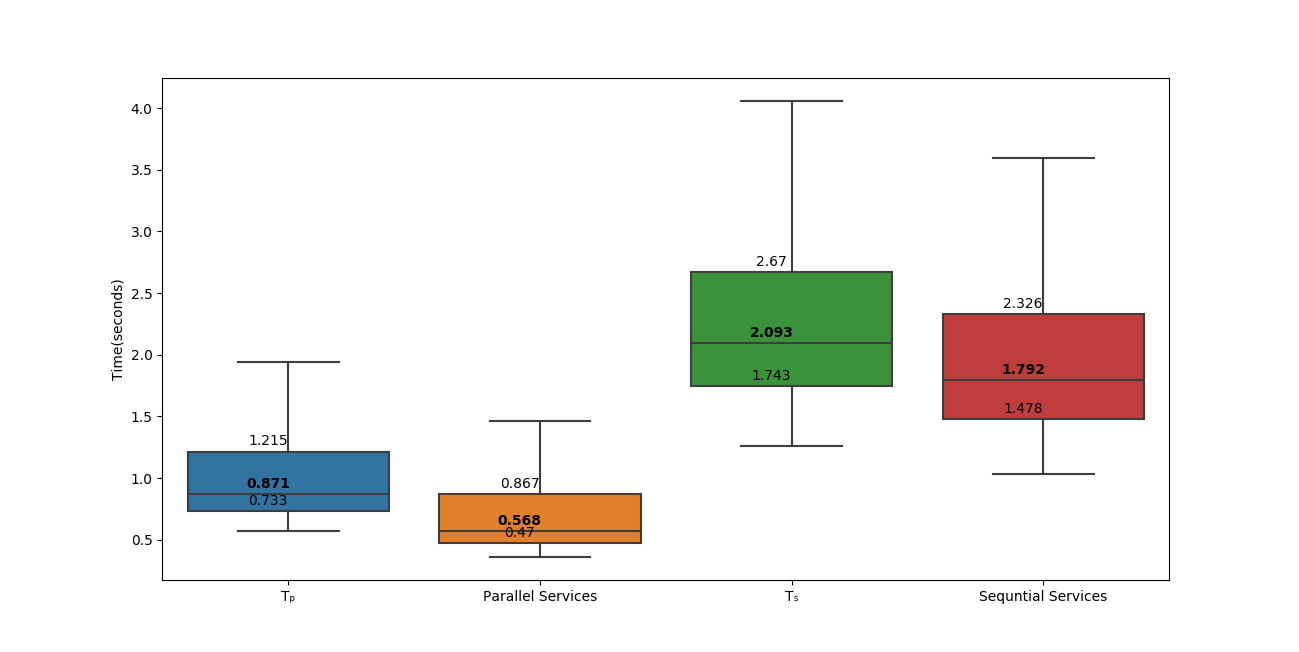}
\caption{Time taken using parallel and sequential calling paradigm}
\label{fig:Time taken using parallel and sequential calling paradigm}
\end{figure}

Calling all independent services one after the other would have been a part of monolithic sequential architecture which is in contrast to micro services oriented parallel architecture proposed. Advantage of parallelized architecture proposed in this study can be best understood by comparing it with sequential architecture. Fig \ref{fig:Time taken using parallel and sequential calling paradigm} shows total time taken (including tika, sectioning classifier, BERT and calling services) for both parallel($T_{p}$) and sequential($T_{s}$) calling style. It also highlight's time taken for multiprocessing and sequential time (calculated by adding time taken by all services). It is interesting to note that proposed parallel architecture has reduced the time taken by independent PaaS calling from 1.792s in sequential to 0.568s, which is more than 3X reduction. This has affected total time taken as $T_{p} = 0.871s$ while $T_{s} = 2.093s$. Numbers quoted are median of the data discussed.  
 
\subsection{Analysis of average response time of CV Parser PaaS}
All PaaS paradigm-based services were deployed on system with Intel(R) Xeon(R) Silver 4114 CPU @ 2.20GHz, 250 GB RAM and 40 core CPU. The performance of CV Parser PaaS on various combinations of concurrency and number of requests are described in table \ref{Performance testing results of CV Parser PaaS(in seconds)}. It is observed that for concurrency as high as 30, the average response time for any number of requests ranging from 10 till 1000, do not exceeds 2.5 seconds, following proposed architecture, which makes the CV Parser PaaS responding in near real time.

For 1000 requests with range of concurrencies varying between 1 and 50, average response time and various percentiles of time taken by requests is mentioned in table \ref{Average and percentile of response time for CV Parser PaaS(in seconds)}. For a concurrency of 1, 3, 5, 10 and 30, the average response time is less than 2 seconds but it increases inclemently as concurrency change from 30 to 50 where average response time and 75th percentile is greater than 2.5 seconds.  
 
\begin{table*}[h]
\caption{Performance testing results of CV Parser PaaS(average response time in seconds)}
\label{Performance testing results of CV Parser PaaS(in seconds)}
\begin{tabularx}{\textwidth}{@{}l*{6}{C}c}
\toprule
Concurrency & 1 & 3 & 5 & 10 & 30\\
No. of Requests & & & & & \\ 
\midrule
10   &  0.689 &  0.729 &  0.824 &  1.253 &  1.268 \\
30   &  0.675 &  0.740 &  0.809 &  1.054 &  2.388 \\
50   &  0.691 &  0.741 &  0.816 &  0.958 &  2.219 \\
100  &  0.691 &  0.726 &  0.781 &  0.912 &  2.006 \\
1000 &  0.686 &  0.728 &  0.778 &  0.863 &  1.847 \\
\bottomrule
\end{tabularx}
\end{table*}

\begin{table*}[h]
\caption{Average and percentile of response time for CV Parser PaaS(in seconds) for 1000 requests}
\label{Average and percentile of response time for CV Parser PaaS(in seconds)}
\begin{tabularx}{\textwidth}{@{}l*{8}{C}c}
\toprule
Concurrency &Average response time  &100th percentile &95th percentile &90th percentile & 75th percentile & 50th percentile & 25th percentile\\ 
\midrule
1    &             0.685739 &   1.013 &  0.763 &  0.742 &  0.711 &  0.680 &  0.656\\
3    &             0.728010 &   2.371 &  0.812 &  0.787 &  0.753 &  0.719 &  0.690\\
5    &             0.778328 &   1.611 &  0.909 &  0.868 &  0.817 &  0.770 &  0.727\\
10   &             0.863276 &   1.676 &  1.060 &  1.008 &  0.930 &  0.857 &  0.776\\
30   &             1.847400 &   3.869 &  2.812 &  2.673 &  2.391 &  1.889 &  1.307\\
50   &             3.145610 &  30.224 &  9.000 &  4.766 &  3.074 &  2.298 &  1.494\\

\bottomrule
\end{tabularx}
\end{table*}

\section{Conclusion}
For bringing fairness in process of hiring and following green HRM practices, deep learning is a promising solution. Proposed parallel architecture and deployment strategy, can help use multiple deep learning models work in production environment, to parse CV document in real time. Empirical study shows that till a concurrency of 30 even with high number of requests the proposed solution could parse and predict using multiple PaaS in less than 2.5 seconds. The high responsiveness of the system can be attributed to selection mechanism for choosing light weight micro web framework. The Proposed set up can be used for developing and deploying similar deep learning applications in production systems.  

\section{Conflict of Interest}
\begin{itemize}
\item Research support was given to both authors working as salaried employees of InfoEdge (India) Ltd. The storage system and CPU-GPU servers of the organization mentioned were used for experiemnting, developing and deploying the architecture and neural network models proposed in the article on the dataset collected of Resumes by the organization.
\item The authors have no conflicts of interest to declare that are relevant to the content of this article.
\item The authors have no financial or proprietary interests in any material/tool discussed in this article.
\end{itemize}

\section*{Acknowledgement}
We would like to thank Pradeep Kumar Singh and Prabin Meitei for many helpful discussions and comments on the proposed architecture along with the members of the Analytics team for training deep learning models,  Quality Assurance team for helping in rigereously doing the performance and stress testing of CV Parser service and making it ready for the bussiness usage.

\bibliographystyle{ieeetr}
\bibliography{CVParser}

\end{document}